\documentclass[10pt,conference]{IEEEtran}
\IEEEoverridecommandlockouts
\usepackage{graphicx}
\usepackage{balance}  
\usepackage{url}
\usepackage[acronym,toc,shortcuts]{glossaries}
\usepackage{epsfig}
\usepackage{amssymb}
\usepackage{amsthm}
\usepackage{multirow}
\usepackage{caption}
\usepackage{subcaption}
\usepackage{siunitx}
\usepackage{soul}
\usepackage{csquotes}
\usepackage{bbm}
\usepackage{dsfont}
\usepackage{xcolor}
\usepackage{adjustbox}
\usepackage{array}
\usepackage{multirow}
\usepackage{fancyhdr}
\usepackage[T1]{fontenc}
\usepackage{makecell}
\usepackage{balance}
\usepackage{algorithm}
\usepackage{algpseudocode}
\usepackage{comment}
\usepackage{pifont} % for checkmarks and cross marks
\usepackage{tikz}
\usetikzlibrary{shapes, arrows.meta, positioning, fit}

\def\BibTeX{{\rm B\kern-.05em{\sc i\kern-.025em b}\kern-.08em
    T\kern-.1667em\lower.7ex\hbox{E}\kern-.125emX}}

\usepackage{pbalance}

\makeatletter
\newcommand{\linebreakand}{%
  \end{@IEEEauthorhalign}
  \hfill\mbox{}\par
  \mbox{}\hfill\begin{@IEEEauthorhalign}
}
\makeatother
\usepackage[table,xcdraw]{xcolor}

\definecolor{lightgray}{gray}{0.9}
\definecolor{lightblue}{rgb}{0.85,0.92,1.0}
\definecolor{lightred}{rgb}{1.0,0.85,0.85}

\definecolor{critical}{rgb}{1.0, 0.8, 0.8}
\definecolor{high}{rgb}{1.0, 0.93, 0.8}
\definecolor{medium}{rgb}{0.88, 0.95, 1.0}

\newacronym{3GPP}{3GPP}{3rd Generation Partnership Project}
\newacronym{5Gr}{5Gr}{5Growth}
\newacronym{5Gr-VS}{5Gr-VS}{5Growth Vertical Slicer}
\newacronym{5Gr-SO}{5Gr-SO}{5Growth Service Orchestrator}
\newacronym{5Gr-RL}{5Gr-RL}{5Growth Resource Layer}
\newacronym{5Gr-VoMS}{5Gr-VoMS}{5Growth Vertical-oriented monitoring system}
\newacronym{AAA}{AAA}{Authentication, Authorization and Accounting}
\newacronym{ACL}{ACL}{Access Control List}
\newacronym{AD-SAL}{AD-SAL}{API-Driven Service Abstraction Layer}
\newacronym{AGV}{AGV}{Automated Guided Vehicle}
\newacronym{AGVc}{AGVc}{AGV controller}
\newacronym{AI}{AI}{Artificial Intelligence}
\newacronym{AP}{AP}{Access Point}
\newacronym{API}{API}{Application Programming Interface}
\newacronym{APN}{APN}{Access Point Name}
\newacronym{AWS}{AWS}{Amazon Web Services}
\newacronym{BCN}{BCN}{Blockchain Network}
\newacronym{BE}{BE}{best-effort}
\newacronym{BET}{BET}{Blind Equal Throughput}
\newacronym{BS}{BS}{Base Station}
\newacronym{BSS}{BSS}{Business Support System}
\newacronym{BSSID}{BSSID}{Basic Service Set Identification}
\newacronym{BTP}{BTP}{Backhaul Transport Provider}
\newacronym{CDR}{CDR}{Call Detail Record}
\newacronym{CELL-ID}{CELL-ID}{cell identification ID}
\newacronym{CI}{CI}{Confidence Interval}
\newacronym{CGN}{CGN}{Carrier Grade NAT}
\newacronym{CLI}{CLI}{Command Line Interface}
\newacronym{CMM}{CMM}{Coordinate Measuring Machine}
\newacronym{CN}{CN}{Core Network}
\newacronym{CPE}{CPE}{Customer Premise(s) Equipment}
\newacronym{CPU}{CPU}{central processing unit}
\newacronym{CQI}{CQI}{channel quality indicator}
\newacronym{CS}{CS}{Circuit Switched}
\newacronym{CSI}{CSI}{channel state information}
\newacronym{CSP}{CSP}{Cloud Service Provider}
\newacronym{CSV}{CSV}{Comma-Separated Values}
\newacronym{D2D}{D2D}{Device-to-Device}
\newacronym{DB}{DB}{Database}
\newacronym{DHCP}{DHCP}{Dynamic Host Configuration Protocol}  
\newacronym{DL}{DL}{Distributed Ledger}
\newacronym{DLT}{DLT}{Distributed Ledger Technology}
\newacronym{DNS}{DNS}{Domain Name System}
\newacronym{DoS}{DoS}{Denial of Service}
\newacronym{DP}{DP}{Dynamic Programming}
\newacronym{DPI}{DPI}{Deep Package Inspection}
\newacronym{DSA}{DSA}{Digital Signature Algorithm}
\newacronym{DTC}{DTC}{Decision Tree Classifier}
\newacronym{DWH}{DWH}{Data Warehouse}
\newacronym{ECDSA}{ECDSA}{Elliptic Curve Digital Signature Algorithm}
\newacronym{eNodeB}{eNodeB}{evolved Node-B}
\newacronym{EC}{EC}{Edge Cloud}
\newacronym{ECC}{ECC}{Elliptic Curve Cryptography}
\newacronym{EPC}{EPC}{Evolved Packet Core}
\newacronym{ENI}{ENI}{Experiential Networked Intelligence}
\newacronym{EPS}{EPS}{Evolved Packet System}
\newacronym{ETSI}{ETSI}{European Telecommunications Standards Institute}
\newacronym{GAS}{GAS}{Grover Adaptive Search}
\newacronym{GBTC}{GBTC}{Gradient Boosting Tree Classifier}
\newacronym{GCP}{GCP}{Google Cloud Platform}
\newacronym{GGSN}{GGSN}{Gateway GPRS Support Node}
\newacronym{GPS}{GPS}{global positioning system}
\newacronym{GRA}{GRA}{Grey relational analysis}
\newacronym{GTP}{GTP}{GPRS Tunneling Protocol}
\newacronym{GNS3}{GNS3}{Graphical Network Simulator-3}
\newacronym{HetNet}{HetNet}{heterogeneous network}
\newacronym{HSS}{HSS}{Home Subscriber Server}
\newacronym{HTTP}{HTTP}{Hypertext Transfer Protocol}
\newacronym{HTTPS}{HTTPS}{Hypertext Transfer Protocol-Secure}
\newacronym{HDFS}{HDFS}{Hadoop Distributed File System}
\newacronym{HiveQL}{HiveQL}{Hive Query language}
\newacronym{IaaS}{IaaS}{Infrastructure as a Service}
\newacronym{ICIC}{ICIC}{inter-cell interference coordination}
\newacronym{ICN}{ICN}{information-centric network}
\newacronym{IEEE}{IEEE}{Institute of Electrical and Electronics Engineers}
\newacronym{IMSI}{IMSI}{International Mobile Subscriber Identity}
\newacronym{IMEI}{IMEI}{International Mobile Station Equipment Identity}
\newacronym{IETF}{IETF}{Internet Engineering Task Force}
\newacronym{InP}{InP}{Infrastructure Provider}
\newacronym{IoT}{IoT}{Internet of Things}
\newacronym{IP}{IP}{Internet Protocol}
\newacronym{IPsec}{IPsec}{Internet Protocol Security}
\newacronym{IT}{IT}{Information Technology}
\newacronym{JSON}{JSON}{JavaScript Object Notation}
\newacronym{KPI}{KPI}{Key Performance Indicator}
\newacronym{KNN}{KNN}{K-Nearest Neighbors }
\newacronym{LAC}{LAC}{location area code}
\newacronym{LAN}{LAN}{local area network}
\newacronym{LCM}{LCM}{Life Cycle Management}
\newacronym{LDA}{LDA}{Latent Dirichlet Allocation}
\newacronym{LDP}{LDP}{Label Distribution Protocol}
\newacronym{LL}{LL}{Logical Link}
\newacronym{LR}{LR}{Logistic Regression}
\newacronym{L2VPN}{L2VPN}{Layer2 Virtual Private Network}
\newacronym{LTE}{LTE}{Long Term Evolution}
\newacronym{MAC}{MAC}{Media Access Control}
\newacronym{MADM}{MADM}{Multiple Attribute Decision Making}
\newacronym{MANO}{MANO}{Management and Orchestration}
\newacronym{MDP}{MDP}{Markov Decision Proces}
\newacronym{MEC}{MEC}{Multi-access Edge Computing}
\newacronym{MEW}{MEW}{multiplicative exponent weighting}
\newacronym{ML}{ML}{Machine Learning}
\newacronym{MME}{MME}{Mobility Management Entity}
\newacronym{MMF}{MMF}{Max-Min Fairness}
\newacronym{MTC}{MTC}{Machine Type Communication}
\newacronym{MNO}{MNO}{Mobile Network Operator}
\newacronym{MO}{MO}{Mobile Operator}
\newacronym{MPI}{MPI}{Message Passing Interface}
\newacronym{MSISDN}{MSISDN}{Mobile Station International Subscriber Directory Number}
\newacronym{MSP}{MSP}{Mobile Service Provider}
\newacronym{MPLS}{MPLS}{Multi Protocol Label Switching}
\newacronym{MT}{MT}{Maximum Throughput}
\newacronym{MVNO}{MVNO}{Mobile Virtual Network Operator}
\newacronym{NAC}{NAC}{Network Access Control}
\newacronym{NAT}{NAT}{Network Address Translation}
\newacronym{NBC}{NBC}{Naive Bayes Classifier}
\newacronym{NBI}{NBI}{Northbound Interface}
\newacronym{NFV}{NFV}{Network Function Virtualization}
\newacronym{NFVO}{NFVO}{Network Function Virtualization Orchestrator}
\newacronym{NIST}{NIST}{National  Institute  of  Standards  and Technology}
\newacronym{NoSQL}{NoSQL}{Not Only SQL}
\newacronym{NS}{NS}{Network Service}
\newacronym{NSD}{NSD}{Network Service Descriptor}
\newacronym{NSIR}{NSIR}{Network Service Instantiation Resource}
\newacronym{NTRU}{NTRU}{N-th degree Truncated polynomial Ring Units}
\newacronym{NWDAF}{NWDAF}{Network Data Analytics Function}
\newacronym{NR}{NR}{New Radio}
\newacronym{QoS}{QoS}{Quality-of-Service}
\newacronym{QoE}{QoE}{Quality-of-Experience}
\newacronym{OAM}{OAM}{Operation, Administration and Management}
\newacronym{ODI}{ODI}{Oracle Data Integrator}
\newacronym{ONAP}{ONAP}{Open Network Automation Platform}
\newacronym{ONOS}{ONOS}{Open Network Operating System}
\newacronym{OS}{OS}{operating system}
\newacronym{OSM}{OSM}{Open Source MANO}
\newacronym{OSS}{OSS}{Operations Support System}
\newacronym{OF}{OF}{OpenFlow}
\newacronym{OTN}{OTN}{Optical Transport Network}
\newacronym{QKD}{QKD}{Quantum Key Distribution}
\newacronym{QUIC}{QUIC}{Quick UDP Internet Connections}
\newacronym{PaaS}{PaaS}{Platform as a Service}
\newacronym{PC}{PC}{Personal Computer}
\newacronym{PCRF}{PCRF}{Policy and Charging Rules Function}
\newacronym{PDN}{PDN}{packet data network}
\newacronym{PDL}{PDL}{Permissioned Distributed Ledger}
\newacronym{PA}{PA}{Placement Algorithm}
\newacronym{PF}{PF}{Proportional Fair}
\newacronym{P-GW}{P-GW}{Packet Gateway}
\newacronym{PKI}{PKI}{Public Key Infrastructure}
\newacronym{PoC}{PoC}{Proof of Concept}
\newacronym{PS}{PS}{Packet Switched}
\newacronym{PNF}{PNF}{Physical Network Function}
\newacronym{PPP}{PPP}{{P}oisson point process}
\newacronym{PHY}{PHY}{physical layer}
\newacronym{PGW}{P-GW}{Packet Data Gateway}
\newacronym{PQC}{PQC}{Post-Quantum Cryptographic}
\newacronym{PoS}{PoS}{Proof-of-Stake}
\newacronym{QRSA}{QRSA}{Quantum Resistant Security Algorithm}
\newacronym{RAM}{RAM}{Random Access Memory}
\newacronym{RAN}{RAN}{Radio Access Network}
\newacronym{REST}{REST}{Representational State Transfer}
\newacronym{RFC}{RFC}{Random Forest Classifier}
\newacronym{RL}{RL}{Resource Layer}
\newacronym{ROE}{ROE}{Resource Orchestration Engine}
\newacronym{RR}{RR}{Round Robin}
\newacronym{RSA}{RSA}{Rivest–Shamir–Adleman}
\newacronym{RSSI}{RSSI}{Received Signal Strength Indicator}
\newacronym{RTT}{RTT}{Round-trip time}
\newacronym{OTT}{OTT}{over-the-top}
\newacronym{OVS}{OVS}{Open Virtual Switch}
\newacronym{OSPF}{OSPF}{Open Shortest Path First}
\newacronym{SaaS}{SaaS}{Software as a Service}
\newacronym{SAC}{SAC}{service area code}
\newacronym{SAW}{SAW}{simple additive weighting}
\newacronym{SDN}{SDN}{Software-Defined Networking}
\newacronym{SLA}{SLA}{Service Level Agreement}
\newacronym{SecGW}{SecGW}{Security Gateway}
\newacronym{SHARING}{SHARING}{Self-organized Heterogeneous Advanced RadIo Networks Generation}
\newacronym{SINR}{SINR}{signal-to-interference-plus-noise ratio}
\newacronym{SCN}{SCN}{small cell network}
\newacronym{SFC}{SFC}{Service Function Chaining}
\newacronym{SGW}{S-GW}{Serving Gateway}
\newacronym{SGSN}{SGCN}{Serving GPRS Support Node}
\newacronym{SSID}{SSID}{Service Set Identification}
\newacronym{SNMP}{SNMP}{Simple Network Management Protocol}
\newacronym{SOE}{SOE}{Service Orchestration Engine}
\newacronym{SSH}{SSH}{Secure Shell}
\newacronym{SVD}{SVD}{singular value decomposition}
\newacronym{SO}{SO}{Service Orchestrator}
\newacronym{SOAP}{SOAP}{Simple Object Access Protocol}
\newacronym{SP}{SP}{Service Provider}
\newacronym{ST}{ST}{Standart Multi-User TOPSIS}
\newacronym{TCP}{TCP}{Transmission Control Protocol}
\newacronym{TEID}{TEID}{tunnel endpoint identifier}
\newacronym{TOPSIS}{TOPSIS}{Total Order Preference By Similarity to the Ideal Solution}
\newacronym{TM}{TM}{Transport Module}
\newacronym{UE}{UE}{user equipment}
\newacronym{UL}{UL}{Upload}
\newacronym{VLAN}{VLAN}{Virtual LAN}
\newacronym{VNF}{VNF}{Virtual Network Function}
\newacronym{VNFD}{VNFD}{VNF Descriptor}
\newacronym{VIM}{VIM}{Virtual Infrastructure Manager}
\newacronym{VL}{VL}{Virtual Link}
\newacronym{VM}{VM}{Virtual Machine}
\newacronym{VoIP}{VoIP}{voice over IP}
\newacronym{VoMS}{VoMS}{Vertical Oriented monitoring system}
\newacronym{WiFi}{WiFi}{Wireless Fidelity}
\newacronym{WLAN}{WLAN}{Wireless Local Area Network}
\newacronym{WMC}{WMC}{weighted Markov chain}
\newacronym{TTI}{TTI}{transmission time interval}
\newacronym{TWAMP}{TWAMP}{Two-Way Active Measurement Protocol}
\newacronym{XDR}{XDR}{Extended Data Record}
\newacronym{E-UTRAN}{E-UTRAN}{Evolved Universal Terrestrial Radio Access Network}

\begin{document}
\title{SEAL: An Open, Auditable, and Fair Data Generation Framework for AI-Native 6G Networks \\
%{\footnotesize \textsuperscript{*}Note: Sub-titles are not captured for https://ieeexplore.ieee.org  and should not be used}
%\thanks{This work partially funded by Spanish MINECO grants TSI-063000-2021-54 and TSI-063000-2021-55 (6G-DAWN), Grant PID2021-126431OB-I00 funded by MCIN/AEI/ 10.13039/501100011033, by “ERDF A way of making Europe” (ANEMONE), Generalitat de Catalunya grant 2021 SGR 00770,  ENSURE-6G project funded by the European Union (Grant ID. 101182933) and CONNECT phase 2  project by Research Ireland (Grant no. 13/RC/2077\_P2).}
}

\author{%
Sunder Ali Khowaja$^\ast$, Kapal Dev$^{\dagger}$, Engin Zeydan$^{\kappa}$, Madhusanka Liyanage$^{\ddagger}$\\
$^{\ast}$CONNECT Centre,School of Computing, Dublin City University, Ireland\\
$^{\dagger}$CONNECT Centre, Munster Technological University, Ireland\\
$^{\kappa}$ Centre Tecnològic de Telecomunicacions de Catalunya (CTTC/CERCA), Castelldefels, Spain\\
$^{\ddagger}$CONNECT Centre, School of Computer Science, University College Dublin, Ireland\\
\protect Email: sunderali.khowaja@dcu.ie,  kapal.dev@mtu.ie, engin.zeydan@cttc.cat, madhusanka@ucd.ie
}

\maketitle

\begin{abstract}

AI-native 6G networks promise to transform the telecom industry by enabling dynamic resource allocation, predictive maintenance, and ultra-reliable low-latency communications across all layers, which are essential for applications such as smart cities, autonomous vehicles, and immersive XR. However, the deployment of 6G systems results in severe data scarcity, hindering the training of efficient AI models. Synthetic data generation is extensively used to fill this gap; however, it introduces challenges related to dataset bias, auditability, and compliance with regulatory frameworks. In this regard, we propose the Synthetic Data Generation with Ethics Audit Loop (SEAL) framework, which extends baseline modular pipelines with an Ethical and Regulatory Compliance by Design (ERCD) module and a Federated Learning (FL) feedback system. The ERCD integrates fairness, bias detection, and standardized audit trails for regulatory mapping, while the FL enables privacy-preserving calibration using aggregated insights from real testbeds to close the reality-simulation gap. Results show that the SEAL framework outperforms existing methods in terms of Frechet Inception Distance, equalized odds, and accuracy. These results validate the framework's ability to generate auditable and bias-mitigated synthetic data for responsible AI-native 6G development. 
\end{abstract}

\begin{IEEEkeywords}
AI-native 6G, synthetic data generation, dataset auditability, data governance, Audit trails.
\end{IEEEkeywords}

% \tableofcontents

\section{Introduction}
Wireless communication networks have evolved since the inception of artificial intelligence; however, fifth-generation (5G) systems have paved the way for seamless integration of AI with communication networks. 5G is currently evolving into AI-native sixth-generation (6G) communication systems, which integrate AI across all layers to enable intelligent, adaptive, and autonomous operations \cite{ref1, ref2}. 
% Unlike previous generations of communication systems, 6G aims to achieve a paradigm in which AI drives network optimization, resource allocation, fault prediction, and enhanced user experience in real time . 
This paradigm shift has been driven by several factors, including exponential growth in data traffic, the proliferation of Internet of Things (IoT) devices, the demand for ultra-reliable low-latency communications (URLLC), and massive machine-type communications (MMTC), that enable applications such as autonomous vehicles, smart cities, and immersive extended reality (XR). Currently, 6G systems cannot be tested in real time, as they are mostly deployed as testbeds or during standardization processes, therefore, the 6G systems heavily on synthetic data generation to train AI models \cite{ref3}. 
% Synthetic data generation can help train AI models for domains such as ray tracing for channel modeling and stochastic processes for traffic patterns, as suggested in \cite{Sionna}. 
For the aforementioned applications, synthetic data generation allows researchers and operators to prototype and validate AI algorithms without relying on limited empirical data. It also facilitates the transition from theoretical designs to practical implementation across diverse scenarios, including urban mobility and interference dynamics, making data generation essential for realizing the full potential of AI-native 6G.

Existing studies highlight the importance of synthetic data generation, but rarely address the auditability of these techniques. In AI-native 6G networks, auditability concerns the trustworthiness and ethical deployment of intelligent systems within telecommunication infrastructure. 
% Synthetic data enables the creation of diverse training sets that replicate real-world vulnerabilities, such as distribution shifts in user behavior or environmental perturbations, which help improve model generalization and robustness \cite{SDG1}. 
The problem with synthetic data generation is that it cannot be determined whether the datasets and resulting AI models are transparent, verifiable, and free from biases unless they are audited. Bias and lack of transparency can result in discriminatory outcomes, such as uneven service quality across demographics \cite{Fairness1, Fairness2}. Therefore, auditing datasets and AI models is essential. Auditability, which includes provenance tracking and compliance mapping to regulatory standards, also fosters trust among stakeholders when implementing applications for high-risk systems under EU AI Act guidelines \cite{DatasheetforDataset}.

% In the context of 6G networks, where AI handles dynamic spectrum access and beamforming in millimeter-wave (mmWave) and terahertz (THz) bands, synthetic datasets reduce the risks associated with data security and privacy concerns under regulations like the General Data Protection Regulation (GDPR) \cite{GDPR}. 

% The auditing mechanisms, including provenance tracking and compliance mapping to standards like the EU AI Act, improve the data practices from being just the technical tools to the socio-technical imperatives. 

% It is also imperative that the auditability should not be conducting just for the sake but rather in an efficient manner. 

It is important to understand that, if not properly audited, synthetic generation could inadvertently propagate errors or spurious correlations, compromising network reliability and security in scenarios involving edge computing for autonomous systems \cite{Security1}. Therefore, integrating these elements is crucial for sustainable 6G ecosystems that balance innovation with accountability. Recently, significant interdisciplinary attention has focused on advancements in AI-native 6G, synthetic data generation, and auditability. 
For AI-native 6G, studies such as \cite{FL2, FL1} examine the integration of machine learning for network orchestration, emphasizing federated learning (FL) to enable privacy-preserving model training in distributed environments, while the study \cite{Agentic} addresses the introduction of Agentic AI techniques in communication systems. Researchers have also explored various methods to improve synthetic data generation, including \cite{Sionna}, which uses ray-tracing simulations to generate channel datasets for 6G physical layer research, and \cite{OpenGym}, which leverages OpenRAN Gym for real-time data synthesis in disaggregated RANs. Auditability in 6G networks has been addressed only through checklists or static methods, as proposed in \cite{Fairness1} and \cite{Fairness2}. These studies introduce fairness tools such as AIF360, which was adopted for 6G bias detection, and datasheets for dataset transparency \cite{DatasheetforDataset}.

% To achieve privacy preservation in 6G simulations, FL has been used extensively to demonstrate privacy-enhanced resource management with a focus on adversarial robustness \cite{FL1, FL2, FL3}.
% These contributions collectively advance simulation-based prototyping and ethical AI integration. \\

Although the advancements proposed in existing works are promising, gaps still persists, particularly in frameworks that integrate synthetic data generation, auditability, and adaptive calibration in AI-native 6G. Many studies on synthetic data generation \cite{Sionna, OpenGym} prioritize realism but overlook ethical safeguards such as causal bias detection or regulatory compliance, which can result in potentially discriminatory models \cite{JMLR:v17:14-518}. Auditability efforts \cite{Fairness1, Fairness2, DatasheetforDataset}, on the other hand, provide metrics but lack integration with dynamic 6G simulations, where distribution shifts can undermine model stability. FL-based approaches \cite{FL2, FL3} address privacy concerns but often ignore the simulation-to-reality gap, resulting in suboptimal performance in real deployments \cite{realitygap}. Furthermore, existing works and architectures treat ethics as a post hoc or secondary check rather than integrating them as a design principle, which limits adoption in regulated telecom sectors\footnote{https://www.nist.gov/itl/ai-risk-management-framework}. 
% Through the proposed work, we aim to bridge the gap by introducing an open data framework with an Ethical and Regulatory Compliance by Design (ERCD) module, along with a closed-loop FL feedback system that incorporates adversarial testing, causal bias scoring, and privacy-preserving calibration to ensure auditable, realistic synthetic data from the outset. 

In this regard, we propose a formalized closed-loop framework for auditable synthetic data in AI-native 6G, integrating Ethical and Regulatory Compliance by Design (ERCD) for ethical compliance and FL for continuous realism enhancement. By addressing challenges such as bias propagation and simulation-to-real-world discrepancies through adversarial suites, causal discovery, audit trails, and federated insights from testbeds, the proposed framework ensures responsible innovation in AI-native 6G\footnote{https://artificialintelligenceact.eu/}. The proposed framework is intended to serve as a foundation for standards bodies to promote safe and lawful AI-native 6G adoption. The specific contributions of this study include:
\begin{itemize}
    \item We propose a layered architecture with ERCD components for regulatory-aligned data pipelines.
    \item We propose a novel FL calibration to dynamically refine simulations and improving fidelity.
    \item We propose method for generating benchmark datasets with enhanced fairness.
    \item We perform the experiments that validate the usability of proposed method on standard hardware.
\end{itemize}
\begin{figure*}[h]
\centering
\includegraphics[width=0.8\linewidth]{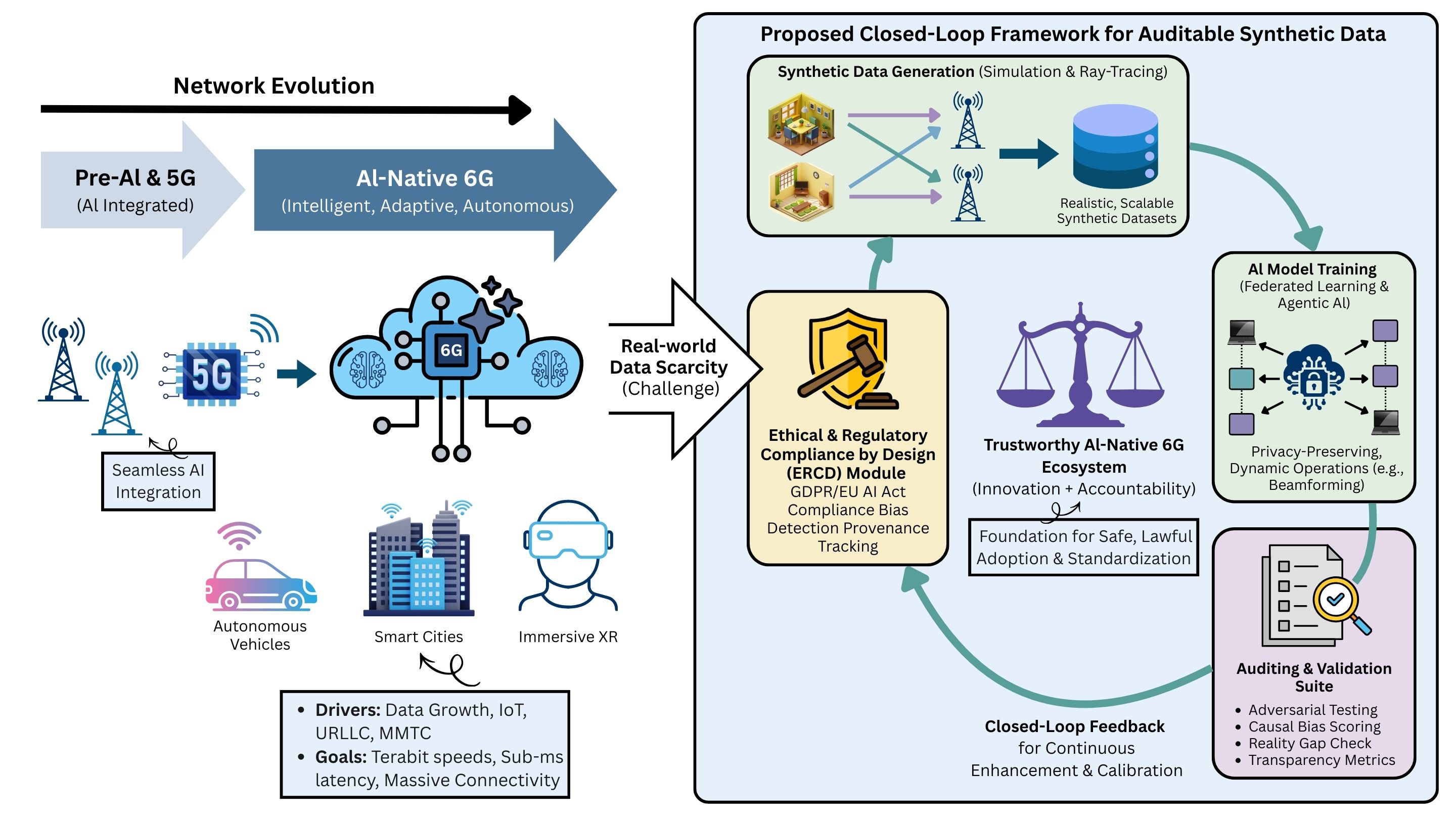}
\caption{The network evolution from 5G to AI-native 6G embedded with auditable synthetic data.}
\end{figure*}
\begin{figure*}[h]
\centering
\includegraphics[width=0.75\linewidth]{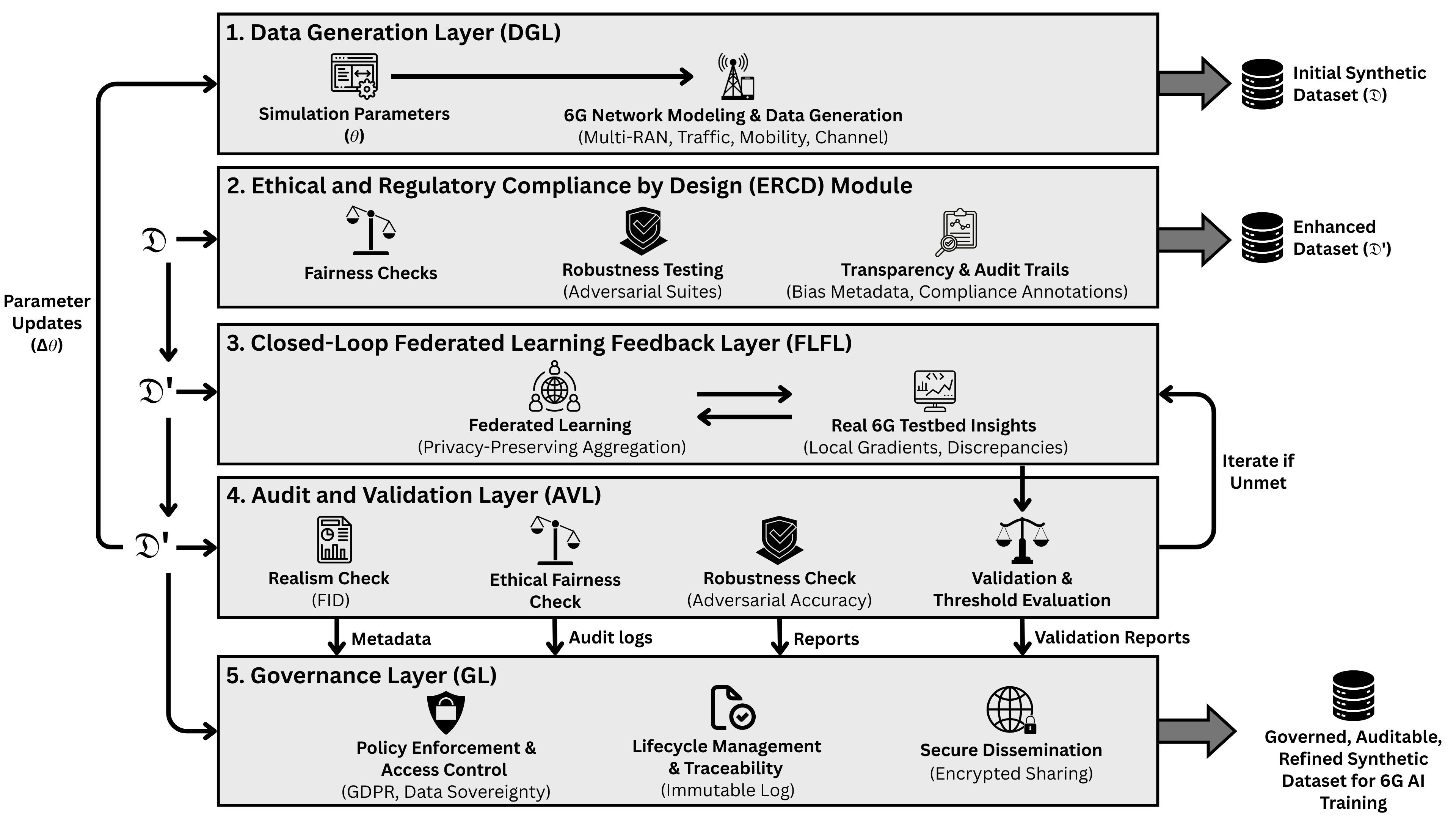}
\caption{The proposed SEAL framework as a layered architecture for auditable and fair synthetic data generation.}
\end{figure*}

\section{Proposed Framework }
In this section, we present the details for the proposed closed-loop framework, i.e. Synthetic data generation with Ethics Audit Loop (SEAL) in AI-native 6G networks. 
% The SEAL framework focuses on simulating realistic 6G scenarios through layered, configurable components. By integrating the ERCD module and a closed-loop FL feedback system, the SEAL framework ensures that synthetic datasets are not only realistic but also ethically sound, auditable, and continuously refined against real-world discrepancies. 
The rationale for this work is to overcome issues concerning data scarcity, ethical safeguards, and the simulation-to-reality gap in AI-native 6G. The SEAL framework is proposed to be method-agnostic that allows flexibility in the choice of specific simulation techniques, bias detection algorithms, and FL aggregation methods.
% The SEAL framework also prioritizes structural cohesion, interoperability, and compliance with standards such as 3GPP TS 28.105 for AI trustworthiness\footnote{https://www.3gpp.org/technologies/ai-ml-management}. \\
The proposed SEAL framework comprises of five layers, i.e. Data Generation Layer, ERCD Module, FL feedback layer, Audit and Validation layer, and Governance layer. The SEAL framework is designed to be an integrated pipeline that processes inputs from simulation parameters to produce auditable and refined synthetic datasets that are suitable for training AI models in the context of AI-native 6G systems \cite{ref1}. Let us denote the simulation parameters with $\theta$, which is fed into the data generation layer (DGL) to generate the initial synthetic dataset $\mathfrak{D}$ while capturing 6G-specific scenarios. The $\mathfrak{D}$ is then fed to the ERCD module that transforms the synthetically generated data into an augmented dataset $\mathfrak{D}'$ by enriching it with test suites, bias metadata, and compliance annotations. To address real-world alignment, the FL feedback layer (FLFL) analyzes the discrepancies between $\mathfrak{D}'$ and aggregated insights from real deployments while computing updates to $\theta$. 
% These updates will refine subsequent generations of $\mathfrak{D}$, indirectly improving future $\mathfrak{D}'$ through iterative recalibration. 
The audit and validation layer (AVL) then assesses the refined $\mathfrak{D}'$ against the quality metrics that can potentially trigger further FL cycles if thresholds are unmet. Lastly, the Governance layer (GL) oversees the entire process while ensuring regulatory compliance across iterations, including FL's privacy-preserving operations. We explain each of the layers in the SEAL framework in subsequent subsections.

% The DGL addresses the data scarcity problem by modeling key network elements such as multi-RANs, traffic patterns, mobility, and channel behavior, hence, ensuring that the generated data supports responsible and efficient AI training for various tasks associated with AI-native 6G networks. 

\subsection{Data Generation Layer}

Synthetic data generation in the SEAL framework is performed at DGL, which emulates the dynamics of AI-native 6G networks. We use a modular, method-agnostic approach for DGL to enable flexible configuration of simulation components while preserving interdependencies across protocol stacks, thus aligning with the need for scalable and realistic synthetic data in emerging 6G ecosystems \cite{Sionna}.  The generation process in DGL can be formalized as $\mathfrak{D} = \mathbb{G}(\theta, \mathfrak{M})$, where $\mathfrak{D}$ denotes the synthetic dataset. The dataset $\mathfrak{D}$ containing $n$ samples can be represented as $\mathfrak{D} = \{d_{1}, d_{2}, \ldots, d_{n}\}$, where each $d_{i}$ is a multidimensional tuple of features such as signal strength, user location, timestamp, and so on. The $\theta$ in the formalization denotes the set of tunable simulation parameters, such as distribution hyperparameters for traffic intensity or mobility speed. Lastly, $\mathfrak{M}$ encapsulates the modular models themselves, which can be selectable functions or processes such as stochastic distributions or physical simulations. These functions define how the network elements are represented and how they interact to produce the data. For example, in traffic modeling, $\mathfrak{M}$ could include a Poisson process to simulate packet arrivals, reflecting realistic workload fluctuations in vertical-aware applications such as IoT or XR. The equation for the Poisson process is $\lambda(t) = \lambda_{0} + \sum_{k=1}^{K} \alpha_{k}sin(2\pi f_{k}t+\phi_{k})$, where $\lambda(t)$ is time-varying arrival rate used to generate packet timestamps in $d_{i}$. In this case, $\theta = \{\lambda_{0}, \alpha_{k}, f_{k}, \phi_{k}|k=1,...,K\}$. The encapsulation $\mathfrak{M}$ is the Poisson framework that applies the $\lambda(t)$ to sample inter-arrival times exponentially that contributes to features in $\mathfrak{D}$.

% The process also ensures that the correlations are preserved when computed across modular outputs. 

Anomalies can also be injected in DGL through probabilistic perturbations, i.e., $\hat{d_{i}} = d_{i} + \epsilon$, where $\epsilon \sim \mathcal{N}(0,\sigma^{2})$ models the noise with $\sigma \in \theta$ and the normal distribution is part of $\mathfrak{M}$. This introduction of anomalies allows the synthetic data generation to represent faults such as interference. The outputs from DGL include metadata for traceability, facilitating integration with downstream layers and supporting ethical, auditable data pipelines in accordance with regulatory requirements \cite{realitygap}. 
% The DGL in SEAL is designed to provide transparency while enabling operators to customize for specific 6G scenarios while maintaining scientific accuracy grounded in stochastic modeling and wireless principles \cite{OpenGym}.

\subsection{Ethical and Regulatory Compliance by Design Module}

The proposed ERCD module in the SEAL framework is designed to integrate ethical and regulatory safeguards directly into the data pipeline. The ERCD module augments the initial dataset $\mathfrak{D}$ from the DGL to produce an enhanced $\mathfrak{D}'$. The enhanced dataset includes checks for fairness, robustness, and transparency. This module is designed to ensure proactive compliance with frameworks such as the EU AI Act and NIST RMF, thereby preventing bias propagation in AI-native 6G applications. Notably, the ERCD module is method-agnostic, meaning that various analytical tools can be used or adopted to align synthetic data generation with real-world representation and the socio-technical trust imperative.

We formally represent the augmentation as $\mathfrak{D}' = \mathfrak{D} \cup T \cup B \cup A$, where $T$ corresponds to adversarial test suites, $B$ represents bias metadata, and $A$ refer to the audit trails. The generation for adversarial test suites is shown in  \eqref{eq:generation}.
\begin{equation}
    T_{r} = \{\phi(d_{i}) | d_{i} \in S \subset \mathfrak{D}\}
    \label{eq:generation}
\end{equation}
where $r$ refers to the regulatory target, e.g., fairness, $S$ is a sampled subset, and $\phi$ refers to a transformation function. The function $\phi(d_{i}) = d_{i}+\epsilon$, where $\epsilon \sim \mathcal{P}(\eta)$. The above expression the $\mathcal{P}$ is a perturbation distribution parameterized by $\eta$, such as Gaussian for noise. The transformation function tests the model's stability under shifts, as suggested in \cite{ref3}. The ERCD module also checks for fairness by resampling demographics: $\hat{d} \sim p(x|g)$, where $g$ is a group attribute (such as urban/rural or another group), and $p$ ensures balanced conditional distributions.

For bias detection, the ERCD module adapts causal scoring by constructing a graph $G = (V, E)$, where vertices $V$ represent the features or parameters and edges $E$ correspond to inference via independence tests. The scoring function used to evaluate spurious correlations is shown in \eqref{eq:scoring}. 

\begin{equation}
    Score = |P(Y|X) - P(Y|X, do(Z))|
\label{eq:scoring}
\end{equation}

In \eqref{eq:scoring}, $Y$ refers to an outcome such as fault prediction, $X$ refers to a protected attribute (e.g., location), and $Z$ corresponds to the confounding variable. The notation $do(Z)$ denotes intervention according to Pearl's do-calculus, which identifies whether generation parameters create discriminatory links\footnote{https://www.nist.gov/itl/ai-risk-management-framework}. We suggest setting the thresholds through bootstrapping. The ERCD module then maps the compliance audit trials through the function shown in \eqref{eq:ERCD}.
\begin{equation}
    A = \{(c_{j}, m_{j}|j=1,...,J\}
    \label{eq:ERCD}
\end{equation}
where $c_{j}$ is a regulatory clause (e.g., EU AI Act Article 10) and $m_{j}$ correspond to a dataset metric such as bias scores. We suggest to perform this structured mapping with JSON to ensure traceability. The outputs of the ERCD module are fed into FL for refinement, which will maintain the coherence by elevating auditability from technical perspective to ethical perspective, which we believe, is desperately needed for 6G's critical infrastructure.

\subsection{Closed-Loop Federated Learning Feedback Layer}

The FLFL task is to calibrate the proposed SEAL framework adaptively by incorporating privacy-preserving insights from real 6G testbeds. This calibration will help refine simulation parameters $\theta$ to close the simulation-reality gap \cite{realitygap}. The FLFL uses the augmented $\mathfrak{D}'$ from ERCD and calculates discrepancies with real data aggregates while updating $\theta$. This process ensures that the subsequent datasets $\mathfrak{D}$ and $\mathfrak{D}'$ achieve greater realism without compromising privacy ethics in 6G distributed environments. Let $CL$ be the number of clients (e.g., testbeds), each with local insights $LI_{CL}$. Discrepancies can be formalized as shown in \eqref{eq:discrepency}.
\begin{equation}
    \delta_{CL} = \frac{1}{m}\sum_{i=1}^{m} \|f_{\theta}(x_{i})-y_{i}\|^2
    \label{eq:discrepency}
\end{equation}
where $f_{\theta}$ is the simulation model predicting states, $x_{i}$ correspond to the inputs from $\mathfrak{D}'$, $y_{i}$ refers to the real observations from $LI_{CL}$, and $m$ refers to the sample size. Local gradients $g_{CL} = \nabla_{\theta}\delta_{CL}$ are aggregated centrally as $g = \sum_{i=1}^{CL} \frac{n_{CL}}{N}g_{CL}$, where the fraction term in the aforementioned formula refers to the FedAvg \cite{FLoriginal}. Updates are performed $\theta_{t+1} = {\theta_{t} - \eta g}$, where $\eta$ is the learning rate. 

The SEAL framework ensures the differential privacy by adding noise, i.e. $v \sim \mathcal{N}(0, \sigma^{2}I)$ to $g_{CL}$, with $\sigma$ calibrated for $(\epsilon, \delta) - DP$, in order to balance utility and protection \cite{FLoriginal, FL1, FL3}. In the context of 6G, parameters are refined through this process, such as in channel variance: $\theta' = \theta_{chan} + \Delta$, where $\Delta$ corresponds to the parameter that minimizes reality-simulation mismatches. The loop is integrated with audit by providing updated $\mathfrak{D}'$ for validation and governance during aggregation (e.g., consent checks). 

\subsection{Audit and Validation Layer}

The AVL in the SEAL framework evaluates the refined dataset $\mathfrak{D}'$ after calibration by FLFL to ensure that $\mathfrak{D}'$ meets realism, ethical, and quality criteria before release. The AVL is responsible for providing quantitative checkpoints, if needed for triggering iterations, and supporting governance through the generation of validation reports. The AVL assesses realism using the Frechet Inception Distance (FID) \cite{FIDmain} with the formulation shown in \eqref{eq:AVL}.  

\begin{equation}
    FID = \|\mu_{real} - \mu_{sim}\|^2 + Tr(\Sigma_{real}+\Sigma_{sim} - 2(\Sigma_{real}\Sigma_{sim})^{1/2})
    \label{eq:AVL}
\end{equation}
where $\mu_{real}, \Sigma_{real}$ are mean and covariance of real features, $\mu_{sim}, \Sigma_{sim}$ of synthetic $\mathfrak{D}'$, and $Tr$ is the trace that measures distributional similarity. We evaluate the ethical fairness using equalized odds \cite{Eqodd} such that $\Delta = |P(\hat{Y} = 1|Y=0, A=0) - P(\hat{Y}=1|Y=1, A=1)| + |P(\hat{Y}=1|Y=1,  A=0) - P(\hat{Y}=1|Y=1, A=1|)$, with $\hat{Y}$ being the predictions, $Y$ being the true labels, and $A$ being the protected attributes, to ensure non-discrimination. We compute adversarial accuracy to validate robustness in AVL as shown in equation 6
\begin{equation}
    Acc = \frac{1}{n}\sum_{i=1}^{n} \mathbb{I}(f(\phi(d_{i}))=f(d_{i}))
\end{equation}
where $\mathbb{I}$ is the indicator, $f$ is a test model, and $\phi$ is from the ERCD suites. The thresholds can be set, for instance $FID<0.1, \Delta<0.05$ to determine pass/fail, which will then signal the FLFL to run the calibration again or not. 

\subsection{Governance Layer}

The GL oversees the entire framework and enforces policies on data sovereignty, lifecycle management, and regulatory alignment. By enforcing these policies, the GL advances a sustainable and collaborative 6G data ecosystem. The GL incorporates trustworthy data practices by integrating previous components, including ERCD's compliance trails, FL's privacy mechanisms, and audit validations, to maintain ethical integrity across iterations. The GL manages access controls and dissemination, following established data governance principles such as those in the GDPR. We formalize the authorization function in the governance as shown in \eqref{eq:authorization}.

\begin{equation}
    Auth(\mathfrak{u}, \mathfrak{D}') = \Lambda_{l=1}^{L} Eval(pm_{l}, \mathfrak{u}, \mathfrak{M})
    \label{eq:authorization}
\end{equation}
where the notation $Auth$ is a boolean function that determines if the user $\mathfrak{u}$ (e.g., a researcher or operator) is granted access to the augmented dataset $\mathfrak{D}'$. The notation $pm_{l}$ refers to the policy mapping such that $PM = \{pm_{l} | l= 1,...,L\}$. The set of policies can be either role-based rules or consent requirements. The notation $\mathfrak{M}$ is the metadata from $\mathfrak{D}'$, the $Eval$ function corresponds to the evaluation of each policy $pm_{l}$ against $\mathfrak{u}$ and $\mathfrak{M}$ that returns true of the compliance is met. Lastly, the operator $\Lambda$ denotes the logical AND operation suggested that the access is granted only if all policies are satisfied simultaneously. The $Auth$ function ensures that multi-criteria checks are performed to verify user credentials and data certifications, preventing unauthorized exposure in shared 6G environments \cite{GDPR}.

In GL, we achieve the traceability through lifecycle state management, i.e. $\mathcal{S}_{t+1} = Trans(\mathcal{S}_{t}, \mathcal{A}_{t})$, where $\mathcal{S}_{t}, \mathcal{A}_{t}$ represent the dataset's state at time $t$ (e.g., "generated", "validated", "archived"), and the action performed (e.g., FL update or audit pass), respectively. The $Trans$ function is the transition function that records changes with timestamps and the actors involved. Lifecycle state management uses provenance metadata from the DGL and ERCD trails to create an immutable audit log, enabling reconstruction of the dataset's history for compliance audits or dispute resolution \cite{auditgovernance}. We can also apply differential privacy in the Governance layer by integrating a privacy budget into the loss function. Lastly, dissemination, that is, the sharing of datasets in GL, is managed through the function $Share(\mathfrak{D}') = Enc(\mathfrak{D}', key)$, indicating that if $Auth$ succeeds and certification from the audit is valid, the data can be shared in encrypted form $Enc$ using the encryption key $key$. Encrypted sharing of the dataset supports interoperability across organizations while enforcing data sovereignty. 

\section{Experimental Setup and Analysis}

To evaluate the proposed SEAL framework, we conducted experiments on a single PC equipped with an NVIDIA RTX 4090 GPU to demonstrate that the proposed framework is feasible for individual researchers without relying on distributed computing resources. We used GPU acceleration for tasks such as dataset generation, causal graph computation, and FL simulations. In this paper, we intentionally used open-source tools and libraries to promote reproducibility and compliance. For example, we used Python as the programming language, PyTorch for tensor operations and model training, NetworkX\footnote{https://networkx.org/en/} for causal discovery graphs, AIF360\footnote{https://aif360.readthedocs.io/en/latest/index.html} for fairness metrics, and Sionna for ray-tracing-based channel simulations \cite{Sionna}. The evaluation was conducted on synthetic data generation using DFL in the context of 6G scenarios.

To generate the synthetic dataset $\mathfrak{D}$, we simulated a 6G network slice with 100 users over a 1 $km^2$ urban area, producing 10,000 samples per run. The features in the synthetic dataset included traffic loads (Poisson-distributed with $\lambda = 5$ packets/second), mobility (random waypoint with speeds of 1–10 m/s), and channels (ray-traced with mmWave frequencies at 28 GHz). We injected anomalies such as distribution shifts, for example, a 20\% traffic surge, to mimic real-world variability. The synthetic dataset $\mathfrak{D}$ was augmented using the ERCD module to obtain $\mathfrak{D}'$ by incorporating adversarial suites (e.g., 20\% perturbed samples via additional noise), bias scoring, and audit trails.

For the FL calibration, we simulated a federated environment with five virtual clients, running ten rounds of FedAvg with differential privacy noise set to 1.0. Real-world insights were emulated by adding noise to the baseline data, specifically 15\% interference, to ensure privacy-preserving updates to the parameters $\theta$. Hyperparameters such as learning rate, batch size, and FL rounds were selected empirically through grid search on a held-out validation set. Selection was based on convergence speed and metric stability. Early stopping was applied if the validation loss did not improve for three epochs. For downstream evaluation, we trained a simple deep neural network with three layers, 128 units, and ReLU activation for resource allocation. The network was trained using the Adam optimizer and cross-entropy loss.

All experiments were repeated five times to assess statistical significance. We used the FID metric to evaluate realism, Equalized Odds (EO) to assess fairness, and accuracy to measure task performance, which was computed post-audit in AVL. The experimental results indicate that the proposed SEAL framework improves dataset quality and downstream model performance in AI-native 6G simulation, while also integrating ethical compliance and FL calibration. The results are shown in Table \ref{tab:my-table}. We compare these results to baselines, including Sionna \cite{Sionna}, OpenRAN Gym \cite{OpenGym}, and AI-driven frameworks \cite{FL2, FLprivacysynth}. We also include the work in  projects such as 6Garrow\footnote{https://www.6gflagship.com/news/towards-ai-native-6g-networks/} in the comparison to provide context for FL and ethics. It should be noted that although we made a direct comparison, the comparisons were adapted due to differing scopes. It should also be noted that we report an estimated value for 6GArrow based on their statements and figures on their project website, which is between Sionna (0.12) and a recent work \cite{sovereignai6gfuture} (0.10). We also want to highlight that although improvements in results are observable, they are constrained by the simulated nature of "real-world" insights and the scale of experiments, thus, they best define early-stage 6G prototyping \cite{ref1} rather than a full-scale development.

\begin{table}[htp!]
\caption{Comparative analysis of FID, EO, and Acc for the proposed method with existing works}
\label{tab:my-table}
\begin{tabular}{|c|l|l|l|}
\hline
\textbf{Method} & \multicolumn{1}{c|}{\textbf{FID}} & \multicolumn{1}{c|}{\textbf{EO}} & \multicolumn{1}{c|}{\textbf{Acc}} \\ \hline
\textbf{Sionna} & 0.12 $\pm$ 0.03 & N/A & 85 $\pm$ 3 \\ \hline
\textbf{OpenRAN Gym} & 0.15 $\pm$ 0.04 & 0.70 $\pm$ 0.05 & 88 $\pm$ 2 \\ \hline
\textbf{{}\cite{FL2}{}} & N/A & 0.78 $\pm$ 0.04 & 95.5 $\pm$ 1 \\ \hline
\textbf{{}\cite{FLprivacysynth}{}} & N/A & 0.82 $\pm$ 0.03 & 90 $\pm$ 2 \\ \hline
\textbf{{}6GArrow{}} & 0.11 $\pm$ 0.03 (est) & 0.80 $\pm$ 0.04 & 91 $\pm$ 2 \\ \hline
\textbf{SEAL} & 0.09 $\pm$ 0.02 & 0.85 $\pm$ 0.03 & 92 $\pm$ 2 \\ \hline
\end{tabular}
\end{table}

The key findings show that the proposed SEAL framework reduces FID by 25\% compared to the uncalibrated baselines, highlighting the effectiveness of FL refinements in improving realism. However, the results are only 10–15\% better compared to Sionna's pure simulations. The proposed SEAL framework improves EO by 20\% while mitigating biases through causal detection. The results show that the SEAL framework outperforms AIF360 standalone \cite{Fairness1} by 12\% but performs similarly to the study \cite{FLprivacysynth}. The SEAL framework reports a task accuracy of 92\%, representing a 10\% gain over baselines that do not use ERCD. However, the proposed work achieves a lower performance score compared to the study \cite{FL2}, primarily due to the privacy noise. Therefore, the proposed SEAL framework represents a modest trade-off in privacy, ethics, and computation. It is also important to note that the strength of the SEAL framework lies in its holistic integration, including the addition of an ethical module to achieve auditability, which is absent in many baselines.

\section{Conclusion}

This paper proposes the SEAL framework for generating auditable synthetic data in AI-native 6G networks, extending baseline simulation pipelines while ensuring compliance with ethics and regulations. We formalize a layered architecture that integrates data generation, ethical augmentation, adaptive calibration, audit validation, and governance, all of which are critical challenges associated with 6G development. The framework also addresses issues related to data scarcity, the simulation-to-reality gap, and proactive compliance with the EU AI Act or NIST RMF. We emphasize that the proposed framework is method-agnostic and therefore flexible in implementation. Experimental results show that the proposed framework improves the FID, which represents the simulation-to-reality gap; fairness, which corresponds to bias; and task accuracy, which demonstrates the ability to produce high-quality datasets. We show that the proposed framework achieves better results compared to existing works.

Although the advantages are clear, we also want to highlight the limitations of the SEAL framework, particularly its reliance on emulated real-world insights. Performance may vary when integrated with actual 6G testbeds or when scaled to larger federations. Therefore, as future work, we plan to extend this study by integrating the SEAL framework with 6G testbeds and increasing the number of devices in the federation to more than 100. Apart from its limitations, we believe that the SEAL framework provides a foundation for the safe, lawful, and ethical evolution of synthetic data generation for AI-native 6G systems. 

\balance
\bibliographystyle{IEEEtran}
\bibliography{References}

\end{document}